# Online User Profiling to Detect Social Bots on Twitter

Maryam Heidari
George Mason University

James H Jr Jones
George Mason University

Ozlem Uzuner
George Mason University

*Abstract*—Social media platforms can expose influential trends in many aspects of everyday life. However, the trends they represent can be contaminated by disinformation. Social bots are one of the significant sources of disinformation in social media. Social bots can pose serious cyber threats to society and public opinion. This research aims to develop machine learning models to detect bots based on the extracted user's profile from a Tweet's text. Online users' profile shows the user's personal information, such as age, gender, education, and personality. In this work, the user's profile is constructed based on the user's online posts. This work's main contribution is three-fold: First, we aim to improve bot detection through machine learning models based on the user's personal information generated by the user's online comments. The similarity of personal information when comparing two online posts makes it difficult to differentiate a bot from a human user. However, in this research, we turn personal information similarity among two online posts as an advantage for the new bot detection model. The new proposed model for bot detection creates user profiles based on personal information such as age, personality, gender, education from user's online posts, and introduces a machine learning model to detect social bots with high prediction accuracy based on personal information. Second, create a new public data set that shows the user's profile for more than 6900 Twitter accounts in the Cresci 2017 [1] data set. All user's profiles are extracted from the online user's posts on Twitter. Third, for the first time, this paper uses a deep contextualized word embedding model, ELMO [2], for social media bot detection task.

## I. INTRODUCTION

Since social media is a place to exchange information, it is essential to differentiate between human posts and comments generated by bots. Bots intentionally spread information on these platforms to create a specific trend that could manipulate public opinion [3]. Social media platforms can be a rich source of user content, and user's personal information can be disclosed in online platforms [4] and can be misused by social bots or Online fake identities to pose serious threats to financial organizations [5]. Bots can distribute fake news in social media and create disinformation in the society . Social media platforms can be the primary source of news and narratives about critical events in the world [6], and bots can significantly affect these events. Bots spread low-quality information and even misleading news, which can be hard to detect based on content. Social media bots can be created to target various audiences [1]. One study identified multiple types of spambots, including promoter bots, URL spam bots, and fake followers [1]. Promoter bots spend several months on promoting specific hashtags to create fake trends or promote specific products [1]. For example, they can promote a sale on Amazon or help the political campaign to win an election [7]. URL spam Bots spread scam URL links by embedding them in retweets they create of legitimate users. Other types of bots include fake followers on social media and fake reviewers of specific products [1]. Many studies have attempted bot detection in recent years. For example, in one study, an unsupervised learning approach is used to detect bots that distribute malicious URL links using URL shortening services [8]. Based on this study, URL sharing bots use constant tweet duplication of legitimate users at a specific time to spread malicious URLs. Their results suggest that about 23 percent of accounts that use URL shortening services are bots. Another popular bot detection service is "Botmeter" [1], which is a supervised learning approach to detect social bots. Botmeter uses metadata related to each twitter account, such as network features, user features, and temporal features, to feed the Random Forest classifier algorithm. Network features show how information diffusion happens among multiple groups of users. User features are user name, screen name, the creation time of account, and geographic location. Temporal features show patterns in a tweet's time generation. The community detection approach in social media [9] is used to detect online activities of a group of online users who share similar ideas. For example, DeBot [10]–[12] is another bot detection service that uses the correlation of activities between different accounts. Application of the Benford's law is used for bot detection by analyzing online behaviors of bots. One of the drawbacks of previous models is that we need different information about each user's account, such as users' features and network features, to differentiate a human account from a bot account. However, in real-world scenarios, we need to detect bot accounts in the early stage of posting comments on social media to prevent the spread of misinformation in online communities. In this work, we improve previous models for social media bot detection by using minimum information about each online user's, which is an online post to detect Social Media bots. Another significant difference between our new model for bot detection and previous models is that we use a deep contextualized word representation method to represent each tweet's text to preserve the online comments' context. We use a different implementation of neural network models for bot detection since the successful application of neural network models in different real-life problems such as health [13], cyber security [4] [14] has been proved by different studies. In this research, For each user, personal information such as Personality, age, gender, and education are extracted based on the user's online comments to create the user's profile. This work detects social bots based on the user's profile and provides a natural language processing approach for social media bot detection. In this work for word representation of each tweet's text, we use the bidirectional LSTM model, which is trained by language models based on a large text corpus. Embedding from language models(ELMO) [2] provides us with a rich representation of each tweet's text [15]. In previous studies by Kudugunta et al. [16], LSTM, and GloVe [17] are used for bot detection; They use the account level approach with a combination of metadata related to each user for bot detection. However, this research uses ELMO, a bidirectional LSTM model for the tweet's text, to create contextualized word representations of tweets and extract the user's profile from the online comments. The new model outperforms the previous bot detection models by creating multiple neural network models on the top of multilayer bidirectional LSTM models to extract different aspects of a tweet's text for social media bot detection. This research also creates a new public data set that provides a user's profile for more than 6900 Twitter accounts in the Cresci data set, and it will be available for public research in various aspects of social media network analysis.

---

[1]https://blog.quantinsti.com/detecting-bots-twitter-botometer/

## II. DATASET

The data set for this research is Cresci 2017 data set [1], [18], which is labeled data set of different types of bots and genuine(human) users. Table I shows different types of social media bots in the Cresci data set. There are five categories of social bots. Social Spam bots #1, which are 1000 automated accounts from the 2014 election in Rome. Social spam bots #2 who are promoter bots that spend several months on Twitter to promote specific hashtags on Twitter. Social spam bots #3 are fake followers who share spam URLs on Amazon to promote their products. Traditional spambots who distribute malicious URLs links or fraudulent job offers on Twitter. Fake followers that bought by Cresci authors from different websites. This research chooses Cresci 2017 data set since it includes different types of social spambots. In this data set, each account has the following attributes: follower-count, friends-count, retweet-count, reply-count, number of hashtags, number of shared URL, tweet text, screen name, and user ID. However, in this work, all features are extracted just based on the user's online posts.

Bot purpose is mostly to distribute misinformation. [19]–[25].

Machine learning models have different applications in health, cyber security, business and social computing research [11], [12], [26], [27], [27]–[45]. In this work, our focus is in the fake news, and we do not detect other types of misinformation. We use COVID-19 data set for fake news detection model. We use transfer learning on COVID-19 tweet's text to create a fake news detection model.

TABLE I
CRESCI 2017 DATASET DESCRIPTION

| user type | tweets | Accounts |
|---|---|---|
| genuine | 2839361 | 3474 |
| social spam bot #1 | 1610034 | 991 |
| social spam bot #2 | 428542 | 3457 |
| social spam bot #3 | 1418557 | 464 |
| traditional spam bot | 145094 | 999 |
| fake followers | 196027 | 3351 |

RT @apelike: SHARING! 72-Hour HERPES / COLD SORES / SHINGLES Treatment DISCREET! Visit Amazing Results! http:/ ...

RT thanks man!  @semox74: http://t.co/GPMdqRmpTS ~

New blog post: \\u0e40\\u0e23\\u0e37\\u0e48\\u0e2d\\u0e07

Fig. 1. Tweet Example

## III. DATA PREPROCESSING

Each tweet's text quality can directly affect the tweet word embedding phase, which is used for different classification models and can have a negative effect on training the model. This work uses 284 human annotators who are provided by Amazon Sagemaker [46] to make an annotation for a new public data set of user's profiles for more than 6900 Twitter accounts who post more than 4 million Tweets. Annotators are used to preprocessing the tweets and create the user's profile by using TextGain[2], which is a text-processing API tool that can be used for different text processing tasks such as information retrieval and semantic extraction. Figure 1 shows an example of unwanted characters such as unknown URL types or any characters which can not be removed from the text by using simple regular expression techniques in the text cleaning phase. These types of characters can cause classification errors in the bot detection model. Based on human annotators reports, the tweet's text still has unknown URL types after cleaning the tweet texts. After further analysis by human annotators, and based on figure 1, we find that about 94% of the tweets that use URL shortening services in this data set also has the word 'blog' in the tweet's text. By adding the word "blog" to a customized regular expression, URLs created by URL shortening services are removed from the tweet's text. On the other hand, the existence of the URL in a user's tweet can be essential to detect bots from human accounts, so we add a new attribute to all twitter accounts with the name of URL, which is the indicator of the number of shared URLs by a Twitter account.

Figure 2 shows the results for gender distribution attributes before the pre-processing phase and figure 3 shows the results for gender attributes after data pre-processing by human annotators. The comparison of figure 2 and figure 3 shows the possible effect of text quality on the classification results. The main reason for the difference between figure 2 and figure 3 is that API provides more accurate results about user's gender when it takes user's tweet's text one by one in comparison with taking all user's online comments at once. In this phase, we pass the user's tweets one by one to the API. For each tweet's text, three human annotators repeat the process and check the validation of the final results. This process is repeated for more than four million tweets in the data set. TextGain provides four different confidence numbers for each gender, age, personality, and education. Human annotators assign personality, age, gender, and education to one twitter accounts based on average confidence numbers of all tweets related to that Twitter account. In the user's profile, Online user's age has two categories: age under 25 years old or age over 25. User's gender can be male or female, user's education: educated and not educated, user's personality: introvert, extrovert.

[2]http://textgain.com/

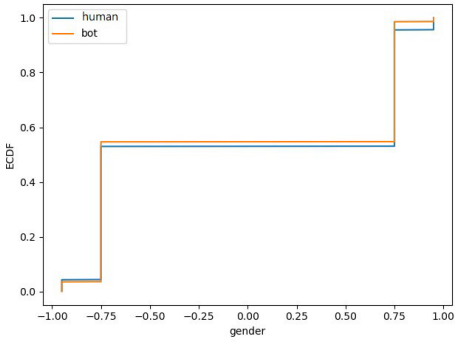

Fig. 2. Gender distribution between bot and human accounts

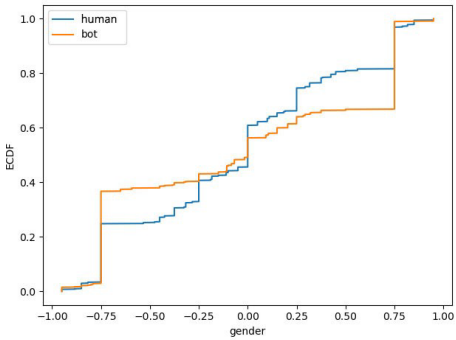

Fig. 3. Improved results for Gender distribution among bot and human accounts after preproseccing

This work creates a customized User interface application to extract the user's profile from a tweet's text. This software helps to customize TextGain API requests based on the text analysis needs and the classification algorithm results. This research will make this application available for the researchers to create an online user's profile for online cyber threat analysis and identity detection on social media research.

## IV. METHODS

The user's profile is added to all Twitter accounts based on the tweet's text in the data preprocessing phase. As shown in figure 3, the distribution of gender is very similar between a bot and a human account. So, attributes with similar values between two classes of bots and humans create obstacles in the bot detection process. For example, It is hard to differentiate between a bot comment and a human comment if both online posts show the same user's profile. However, in the continue, we propose a new bot detection model that turns this obstacle to the advantage for bot detection technique and maximizes the utilization of similar features between two class labels(bot, Human) to improve the bot detection algorithm's prediction accuracy. The new model uses the user's profile to improve prediction accuracy in social media bot detection algorithms.

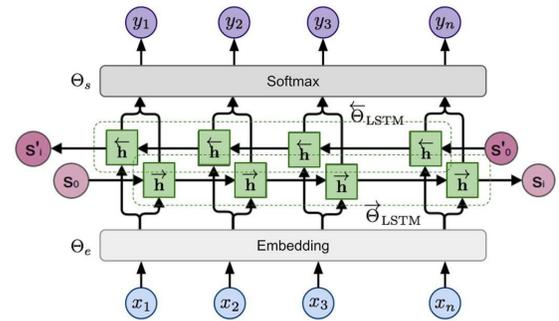

Fig. 4. ELMo[3]

This research uses both Glove(Global Vectors) and ELMO(Embedding from the language model) for word embedding of the tweet's texts to have a complete representation of each tweet's text for input space of bot detection model. Figure 4 shows the Elmo base model that is used in this work. As can be seen, $(x_1, ..., x_n)$, shows the sequence of $n$ tokens in the tweet's text. The bidirectional multi-layer LSTM uses the history of backward and forward tokens for each target token in a tweet's text. The history of backward tokens is calculated by:

$$p(x_1, ..., x_n) = \prod_{i=1}^{n} p(x_i | x_1, ..., x_{i-1}) \quad (1)$$

The history of forward tokens is calculated by:

$$p(x_1, ..., x_n) = \prod_{i=1}^{n} p(x_i | x_{i+1}, ..., x_n) \quad (2)$$

prediction in both directions by multi-layer LSTM can be modeled by hidden states $\overrightarrow{h}_{i,l}$ and $\overleftarrow{h}_{i,l}$ for the input token

bilities of token existence in the tweet's text. We fined tuned

[3]https://www.topbots.com/generalized-language-models-cove-elmo/

the ELMO model[4] to minimize the negative log liklihood in both directions of LSTM model:

$$L = -\sum_{i=1}^{n} \log p(x_i | x_1, ..., x_{i-1}; \Theta_e, \overrightarrow{\Theta}_{LSTM}, \Theta_s) + \log p(x_i | x_{i+1}, ..., x_n; \Theta_e, \overleftarrow{\Theta}_{LSTM}, \Theta_s) \quad (3)$$

In the new social bot detection model, the Elmo model can capture the context of the tweet's text.

Also, the statistics of the global English corpus for each token in a tweet's text can be represented by GloVe in the word embedding phase.

In this work, at all experiments, Tweet's text and original Crescie 2017 data set features are considered an input space for all classification models. First, we examine the effect of adding the user's profile to the input pace of Neural network and logistic regression in the bot detection model. User's age and user's gender have similar distribution between bots and human accounts. Based on data labeling in the data preprocessing phase, each online user has a gender attribute that can be categorized as male or female and have age attributes that can be categorized as online users under age 25 or over age 25. In this section, we examine the effect of similar attributes among online users on Twitter in the bot detection model. Table II shows the results for logistic regression when the user's gender is a new feature for bot detection. Table III shows logistic regression when the user's age is new input for bot detection model.

TABLE II
LOGISTICS REGRESSION BASED ON USER'S GENDER

| Logistic Regression | Accuracy | f1 Score |
|---|---|---|
| without gender | 0.783 | 0.774 |
| with gender | 0.794 | 0.791 |

TABLE III
LOGISTICS REGRESSION BASED ON USER'S AGE

| Logistic Regression | Accuracy | f1 Score |
|---|---|---|
| with age | 0.793 | 0.803 |
| without age | 0.783 | 0.774 |

The prediction accuracy of the model by adding the user's gender is 79%, and adding the user's age is 78%. Comparison of Table II and Table III shows that the user's gender in compared with user's age have more positive effect in prediction accuracy of logistic regression model in bot detection task. However, adding a user's age and gender improve bot detection accuracy by 1%, but we need more improvement in the bot detection model. So there is a need to implement another classification model to capture the real effect of the user's profile in the bot detection model.

Then we examine the effect of the user's age and user's gender in prediction accuracy of the Neural network model. Two layers of feed-forward Neural network are designed for bot detection. Figure 5 shows the FFNN model, for bot classification. Table IV shows the results of the FFNN model based on the user's age. Table V shows the results by adding the user's gender in the neural network model. Adding the user's age and user's gender as a new feature in the single neural network's input space can not improve the prediction accuracy significantly. However, the Neural network model with an accuracy of 82% outperformed the logistic regression with an accuracy of 79% in bot detection by using the user's gender. However, no significant difference in prediction accuracy can be seen by adding the user's profile compared with when we do not add the user's profile.

[4]https://github.com/allenai/allennlp

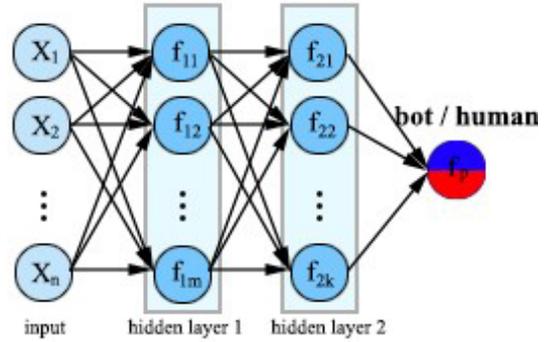

Fig. 5. FFNN

TABLE IV
FFNN RESULTS BASED ON USER'S AGE AFTER

| Algorithm | Accuracy | f1 score |
|---|---|---|
| FFNN with age | 0.808 | 0.805 |
| FFNN without age | 0.807 | 0.803 |

So in this work, the goal is to design a machine learning model that can use the user's profile, which is very similar between class labels(bot and Human) and can detect bots with high prediction accuracy. This work introduces a model that can detect bot accounts, even if two accounts have a similar distribution in age, personality, education, and gender.

Figure 6 shows the new proposed bot detection model based on the user's age. To improve the bot detection model's prediction accuracy, by using an online user's age, we design two different neural networks, one neural network to train data for users over age 25 and another neural network to train users under age 25. As can be seen in Table VI by implementing different neural networks for two age categories of online users, the prediction accuracy of the bot detection model improved by more than 3% in compare with Table IV when we train data related to online users regardless of user's age category and just with one neural network model.

TABLE V
FFNN RESULTS WITH RESPECT TO USER'S GENDER

| Algorithm | Accuracy | f1 score |
|---|---|---|
| FFNN with gender | 0.826 | 0.830 |
| FFNN without gender | 0.824 | 0.829 |

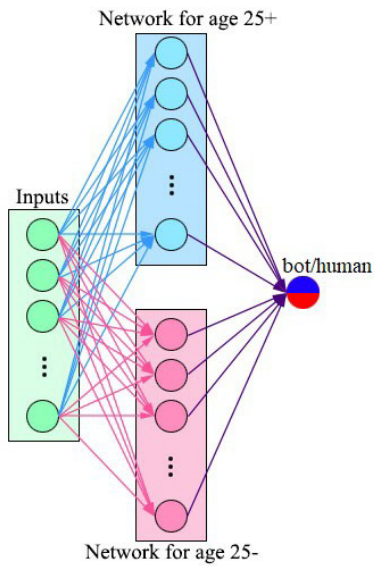

Fig. 6. Multiple Classification for age

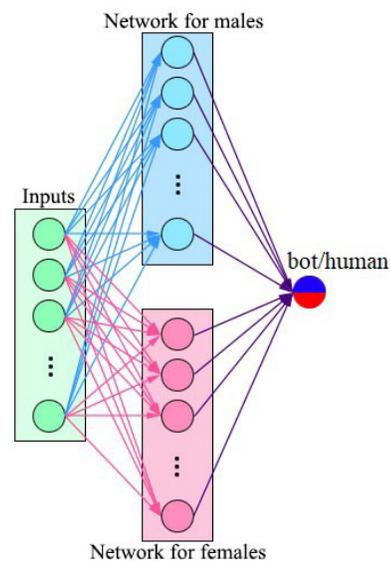

Fig. 7. Multiple Classification for gender

TABLE VI
MULTIPLE CLASSIFICATION RESULTS BASED ON USER'S AGE AND GENDER

| Algorithm | Accuracy | f1 Score |
|---|---|---|
| Multiple Classification with FFNN for Age | 0.846 | 0.838 |
| Multiple Classification with FFNN for Gender | 0.860 | 0.868 |
| The new proposed bot detection model in Figure 10 | 0.882 | 0.868 |

We repeat the same experiment for the user's gender as well. Figure 7 shows the proposed model to detect bot based on the user's gender. One neural network is trained based on female users, and one neural network is trained based on male users. In the female neural network, we have just female accounts, and it is the same in the male network. Table VI shows the results of the multiple classifications of male and female accounts for bot detection. As can be seen, the result shows a nearly 4% improvement in bot prediction accuracy compared with Table V when training male and female users data with one single neural network. Comparison between Table IV, Table V and Table VI shows that to detect bot accounts based on similar user's profile features such as age, gender, personality, and education, we need to implement and train multiple neural networks based on different categories of online user's age, gender, personality, and education. The multiple classification methods significantly affect the bot detection model's prediction accuracy, mainly when the input space of the classification model includes similar features among two class labels(bot, Human). Age, personality, education, gender have similar values among both bot and human accounts; however, with multiple classification models in this work and training the neural network separately for each attribute in the user's profile, we can solve this problem. In our new model, multiple classification models are designed for Personality and Education, as same as age and gender. In the next section, the new bot detection model will be explained.

## V. SAMPLE OF TWEETS AND ERROR ANALYSIS

This section shows a sample of tweets classified correctly and incorrectly by Logistic regression and multiple classification techniques. Red color shows tweets classified by logistic regression, and green color shows tweets classified by new proposed multiple classifications for bot detection based on the user's profile. Figure 8 shows the True positive samples and figure 9 shows false-positive samples for both multiple classification and logistic regression. Tweet classification results of each algorithm are compared side by side. Based on Figure 8, One interpretation of this figures is that for the tweets which have retweet sentiment such as RT@, which can be the indicator of retweet count, the logistic regression has better performance, and the number of tweets which includes @RT word can be seen more in True positive file for logistic regression.

On the other hand, in the false-positive sample for the multiple neural network classification, we can see more tweets, including @RT. The string "RT@" mostly shows retweeting of the specific tweets and can be extracted by simple string matching from a tweet's text. However, features such as age, gender, personality, and education can not be extracted directly from the tweets and have more semantic meaning than simple sentiment. So if the feature has a simple sentiment characteristic rather than a semantic meaning, the logistic regression can capture it better than a neural network model for bot detection. If the feature is distinctive between Bot and human, the logistic regression can capture it better than a neural network. However, if the tweet's text data becomes huge, and the feature becomes complicated semantically, the neural network's multiple classifications outperformed logistic regression. If features can not be observed by only looking at the tweet's text, the neural network's multiple classifications can achieve higher prediction accuracy for the bot detection model. Also, if the features are not very distinctive between humans and bots, like the user's profile in this work, the multiple classifications of the neural network provide higher prediction accuracy than when we train all data by using a single neural network model for bot detection. Based on our knowledge, it could be one reason why multiple classifications outperformed logistic regression and simple FFNN for bot detection in this work since the user's profile attributes can not be extracted from the text by simple sentiment analysis of tweet's text and they have semantic meaning.

Fig. 8. True positive(correctly classified tweets), red color is logistic Regression , green Multiple Classification Neural Network

Fig. 9. False positive(incorrectly classified tweets), red color is logistic regression , green Multiple classifications neural network

## VI. NEW BOT DETECTION MODEL

Figure 10 shows Our proposed model for social media bot detection. As can be seen, the first phase of the new model shows is word embedding of the tweet's text. In this work, GLOVE and ELMO play an important role in the word embedding section, since without complete semantic representation of each tweet, adding the user's profile such as age, gender, education, and personality for input space of the classification model is not effective. The bidirectional LSTM architecture of ELMO is one of the essential components in a new bot detection model. In the second phase, eight different neural networks are trained based on different features in the user's profile. Based on what is discussed in the method section, training data in the neural networks should be in multiple steps, and implementing multiple classifications is key to improving bot prediction accuracy by using the user's profile in the new model.

The last phase in Figure 10 of our new bot detection model is implementing a final classification model that can integrate the final results of all neural network models in the second phase. So the user's profile, which shows age, gender, personality, and education of a user, is the input feature's set for the final classifier in the new bot detection model. We examine different classifiers as a possible candidate for the final classifier. The new model of bot detection needs several experiments to choose the best classifier for the final phase. Table VII shows the results of the prediction accuracy of the bot detection model based on the user's profile. It can be seen that adding the neural network model on the top of all FFNN models in the previous phase outperformed other classifiers in bot prediction accuracy. The results show that the FFNN model with 97% provides the best results for the bot detection model in the last layer of our new model compared with other classifiers. The Random forest provides high prediction accuracy very close to Feedforward neural network. In Table VI, the new model uses age, gender, personality, and education based on the tweet's text. The results show that the new model increases the prediction accuracy by more than 3% compared with previous models and can achieve more than 88% accuracy on more than four million tweets in the Cresci data set. In the next section, we evaluate our new bot detection model compared with previous bot detection techniques based on the same test and training data.

TABLE VII
COMPARE THE CLASSIFIERS

| System | F1-Score | Accuracy | AUC/ROC |
|---|---|---|---|
| Random Forest | 0.969 | 0.962 | 0.889 |
| Logistic Regression | 0.861 | 0.862 | 0.858 |
| AdaBoost Classifier | 0.954 | 0.959 | 0.942 |
| FFNN | 0.970 | 0.971 | 0.972 |
| SGD Classifier | 0.968 | 0.970 | 0.961 |

## VII. PERFORMANCE EVALUATION

To evaluate our new model, We compare our new bot detection model with previous bot detection techniques in table VIII. The new model performance is examined based on the same training and test data set as previous bot detection techniques. The test set #1 is the mix of human and bot accounts in political campaigns, and test set #2 is a combination of human and spambots accounts in Amazon. The new model outperformed previous bot detection models by achieving nearly 94% prediction accuracy in bot detection in both test#1 and test set #2. The new proposed model's main advantage is detecting bot accounts when dealing with a massive amount of social media accounts with very similar characteristics in their user profile. The new proposed model can detect bot and human posts if they have the same age, education, gender, and personality. Also, using ELMO and Glove model for word embedding of tweet's text provides a deep contextualized representation of each tweet's text, which is essential in the bot detection model's prediction accuracy based on the user's profile.

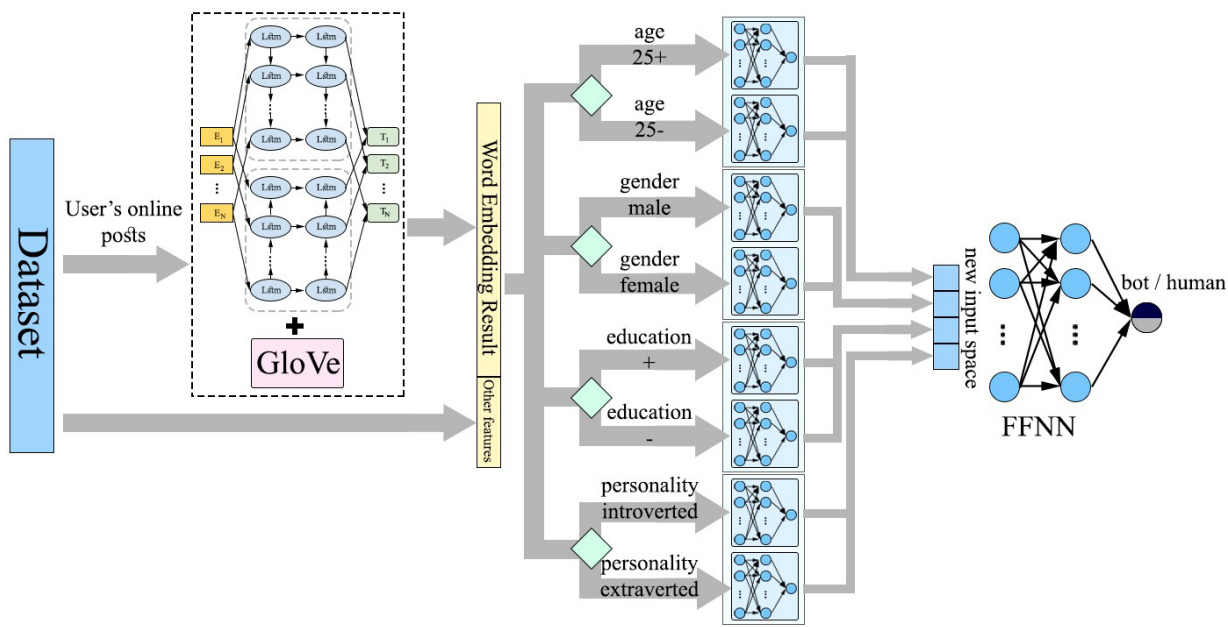

Fig. 10. Our proposed model for social bot detection

TABLE VIII
PERFORMANCE COMPARISON BETWEEN NEW BOT DETECTION MODEL AND PREVIOUS BOT DETECTION TECHNIQUES [47]

| technique | type | Accuracy | F-Measure | MCC |
|---|---|---|---|---|
| test set #1 | | | | |
| Davis et al. [48] | supervised | 0.734 | 0.288 | 0.174 |
| C. Yang et al. [49] | supervised | 0.506 | 0.261 | 0.043 |
| Miller et al. [50] | unsupervised | 0.526 | 0.435 | 0.059 |
| Ahmed et al. [51] | unsupervised | 0.943 | 0.944 | 0.886 |
| Cresci et al. [47] | unsupervised | 0.976 | **0.977** | 0.952 |
| **Our new model** | supervised | **0.981** | 0.976 | **0.962** |
| test set #2 | | | | |
| Davis et al. [48] | supervised | 0.922 | 0.761 | 0.738 |
| C. Yang et al. [49] | supervised | 0.629 | 0.524 | 0.287 |
| Miller et al. [50] | unsupervised | 0.481 | 0.370 | -0.043 |
| Ahmed et al. [51] | unsupervised | 0.923 | 0.923 | 0.847 |
| Cresci et al. [47] | unsupervised | 0.929 | 0.923 | 0.867 |
| **Our new model** | supervised | **0.946** | **0.941** | **0.890** |

## VIII. CONCLUSION

The main contribution of this work can be summarized in four points. First, this research creates a public data set of online users' profiles for more than 6900 twitter accounts. This data set will be available for public research on social media and cyber threat analysis research. Second, using the user's profile, which has similar values between a bot and human accounts, detects bots in social media. The new proposed model in this work can detect bots accounts from human accounts, even if both accounts have the same user profile(education, Gender, Personality, age) are the same. Third, introducing a new bot detection model that can outperform previous bot detection techniques. Fourth, the new bot detection model uses a contextualized representation of each tweet by using ELMO and GLOVE in the word embedding phase, which is essential in achieving the high prediction accuracy in this model.